\documentclass{article}

\usepackage{PRIMEarxiv}

\usepackage[utf8]{inputenc} % allow utf-8 input
\usepackage[T1]{fontenc}    % use 8-bit T1 fonts
\usepackage{hyperref}       % hyperlinks
\usepackage{url}            % simple URL typesetting
\usepackage{booktabs}       % professional-quality tables
\usepackage{amsfonts}       % blackboard math symbols
\usepackage{nicefrac}       % compact symbols for 1/2, etc.
\usepackage{microtype}      % microtypography
\usepackage{lipsum}
\usepackage{fancyhdr}       % header
\usepackage{graphicx}       % graphics
\graphicspath{{media/}}     % organize your images and other figures under media/ folder

\title{Marking anything: Application of point cloud in extracting video target features
}

\author{
  Xiangchun Xu \\
  \texttt{\{xxc18\}@tsinghua.org.cn} \\
}

\begin{document}
\maketitle

\begin{abstract}
Extracting retrievable features from video is of great significance for structured video database construction, video copyright protection and fake video rumor refutation. Inspired by point cloud data processing, this paper proposes a method for marking anything (MA) in the video, which can extract the contour features of any target in the video and convert it into a feature vector with a length of 256 that can be retrieved. The algorithm uses YOLO-v8 algorithm, multi-object tracking algorithm and PointNet++ to extract contour of the video detection target to form spatial point cloud data. Then extract the point cloud feature vector and use it as the retrievable feature of the video detection target. In order to verify the effectiveness and robustness of contour feature, some datasets are crawled from Dou Yin and Kinetics-700 dataset as experimental data. For Dou Yin's homogenized videos, the proposed contour features achieve retrieval accuracy higher than 97\% in Top1 return mode. For videos from Kinetics 700, the contour feature also showed good robustness for partial clip mode video tracing.
\end{abstract}

% keywords can be removed

\section{Introduction}

\begin{figure}
  \centering
  \includegraphics{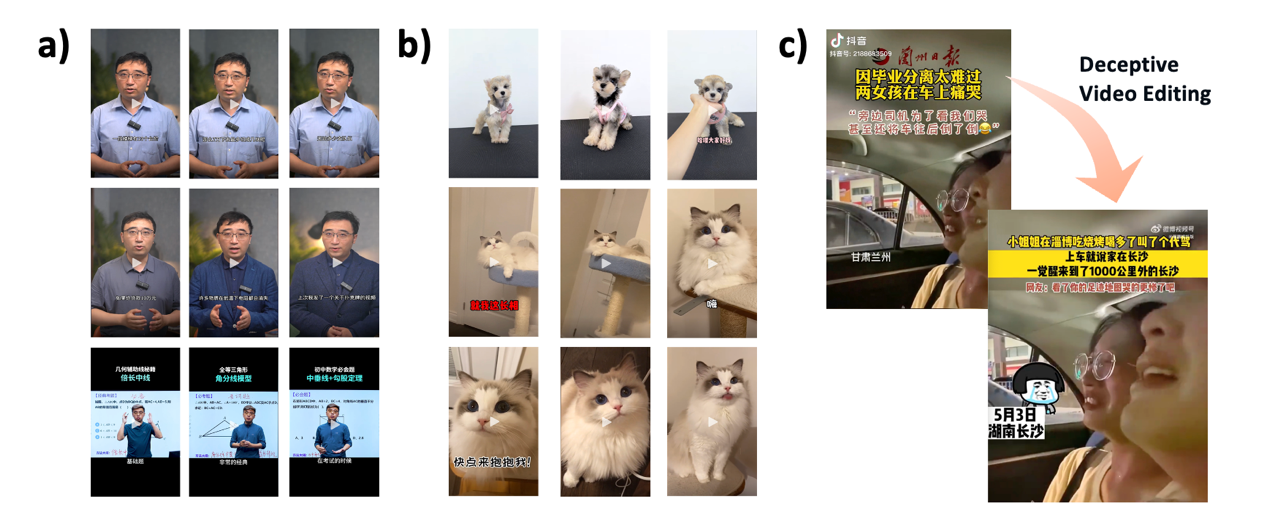}
  \caption{Homogenized video and deceptive video editing.}
  \label{fig:fig1}
\end{figure}

Designing retrievable video features is necessary. With the development of the mobile Internet, anyone can publish videos on streaming media platforms. Massive videos bring higher management requirements. As shown in Fig.\ref{fig:fig1} a)-b), there are many videos with homogeneous content on the Internet at present. Constructing unique ID information for these videos is beneficial to video creators and streaming media Operators. Moreover, the current short video platform is full of many rumored videos, the most classic way of spreading rumors is shown in Fig.\ref{fig:fig1} c), which is to make misleading edits to old videos and replace titles in order to attract attention. Therefore, videos with homogeneous content can be retrieved based on video content extraction features, which is convenient for creators and operation platforms to manage videos. It is also possible to trace the source of false videos as a basis for quickly dispelling rumors.

According to the retrieval mode, the retrieval can be divided into video retrieval based on video, image, text\cite{spolaor2020systematic}. Quite a lot of research has been done on text and image retrieval in academia and industry. Whether it is by matching the voice and subtitles of the video, or by matching the key frames of the video, the existing database technology can meet the precise retrieval of the target video. However, there are few related literature reports on video retrieval by video, or in other words, video retrieval based on video content. The main reason is that compared with words and graphics, videos not only contain rich spatial information, but also rich temporal information. At the same time, for better fluency, the frame rate is also constantly increasing with the advancement of photography technology. Therefore, how to effectively and robustly encode feature vectors for excessively redundant video information is a key step in realizing video retrieval.

At present, the management of many video files is usually to store the video in the file system, and the retrieval of the target video is generally to retrieve the additional information of the video, such as the description of the video, the storage time of the video, and the cover of the video. These types of retrieval method are difficult to achieve accurate video retrieval, because the video additional information is not a direct description of the content of the video itself. In many application scenarios, such attachment information is often missing or not representative. With the development of computer vision technology, there are also many documents or patents that extract video key frames and use neural network to encode key frames to obtain feature vectors, which are used as the basis for video retrieval. Although this method has a small amount of calculation, it does not require too much storage content. However, this method does not make full use of all the information of the video and loses a huge amount of information in the time dimension, therefore, its accuracy often falls short of the ideal level.

With the update of computer hardware, the computing power of current computers has been greatly improved compared with the past. With this advantage, it is possible to use related methods of visual deep learning to analyze video content frame by frame, and to realize vectorized feature extraction of videos based on video content.

Inspired by laser point cloud data processing technology and visual algorithms, this paper proposes an algorithm for extracting feature vectors based on video content. The algorithm first uses the YOLO-v8 model to realize the semantic segmentation of the video frame by frame, so as to obtain the mask information of the detection target; then realizes the tracking of the motion trajectory of the detection target on the video time axis through the multi-object tracking algorithm; Based on these data, the spatial point cloud representation of the detection target is obtained through the boundary extraction technology; finally, the feature encoding of the detection target point cloud data is realized through the PointNet++ network, and this is used as the retrieval basis of the point cloud data.

The first chapter of this paper introduces the current technical solutions and problems faced by video retrieval technology, and briefly introduces the algorithm for extracting video feature vectors based on video content proposed in this paper; the second chapter introduces the visual algorithm and The related work of point cloud algorithm; the third chapter introduces the algorithm framework and training method proposed in this paper in detail; the fourth chapter introduces the validity and robustness verification of the proposed algorithm on real-world data.

\section{Related Work}
\label{sec:headings}
\subsection{MOT algorithm}
The video multi-object tracking algorithm (MOT) is a classic task in computer vision. The main task is to track multiple detection targets in the video and output the position and inter-frame motion trajectory of all detection targets in the video frame. The main processing flow of this type of algorithm for video is divided into two steps, which are intra-frame semantic segmentation and inter-frame motion vector prediction.

Intra-frame semantic segmentation refers to the use of algorithms to determine the pixel position of the detection target frame by frame. Among them, the semantic segmentation model based on YOLO or transformer mechanism is a common algorithm\cite{redmon_you_2016,jiang_review_2022}. Among of them, the YOLO model is a target recognition model based on convolutional neural networks, which can realize end-to-end target recognition of images. Specifically, the algorithm transforms the target recognition classification task into a regression problem and realizes target detection by predicting the bounding box position and target category for each cell in the image. After updating in recent years, the YOLO model has made great progress in detection speed and accuracy. Currently, the YOLO-v8 model has been released. 

Inter-frame tracking refers to predicting the position of a detection target in the current frame in the next frame. According to the number of moving targets to be tracked at one time, inter-frame target tracking is generally divided into single target tracking and multi-object tracking (MOT). The multi-object tracking algorithm can track and detect multiple objects in the video at one time, and can process data efficiently, so it is a commonly used algorithm. SORT is a simple multi-object tracking algorithm that can be processed in real time\cite{wojke_simple_2017,du_strongsort_2023}. It uses the Kalman filter to predict the position of the object and uses the Hungarian algorithm to match the detection target in the current frame with the detection target in the previous frame. At the same time, in order to improve the accuracy of matching, the SORT algorithm also uses the IOU (Intersection over Union) algorithm to measure the degree of overlap between two objects. The SORT algorithm is fast and can calculate and track in real time, but it is more suitable for static shots, and the tracking effect for dynamic shots is not idea. The DeepSORT algorithm is further improved based on the SORT algorithm\cite{pereira_sort_2022}. It uses the convolutional neural network to calculate the characteristics of the detection target and calculates the similarity of the detection target between frames through the cosine distance. At the same time, the model introduces an appearance model to deal with object occlusion and appearance changes, which further improves the robustness of the tracking model. DeepSORT also further enhances the ability to detect objects in dynamic shots.

The OC-SORT model also requires two processes of target detection and tracking, but based on the previous model, the target center tracking mode is added, which calculates the center of the target as the tracking method instead of tracking the outline of the target\cite{cao_observation-centric_2023}. This method further improves the robustness of the target tracking algorithm and reduces the probability of target loss due to occlusion or deformation of the target. The Deep-OC-SORT algorithm is further improved based on the OC-SORT model to obtain higher tracking accuracy and robustness\cite{maggiolino_deep_2023}. The difference from the former is that the model chooses to use the CNN model to extract the features of the detected target as the basis for target matching and tracking.

\subsection{Point cloud algorithm}
In real life, point cloud data is a common form of data representation in radar signal processing. The processing methods for point cloud data are usually acquisition, filtering, down sampling, segmentation, feature extraction, registration, etc\cite{aldoma_tutorial_2012,li_comprehensive_2022}. Generally, different algorithm combinations are selected according to the needs of the task\cite{rusu_fast_2010}. For common point cloud data processing tasks, including but not limited to point cloud recognition, segmentation and clustering, etc., traditional algorithms generally use the extracted point cloud features as the pre-input of these tasks. For example, in point cloud recognition, scholars often use descriptors that characterize the shape of point clouds as the input feature vectors of the model. According to the scale of feature extraction, point cloud descriptors are generally divided into local descriptors and global descriptors\cite{aldoma_global_2012,lu_recognizing_2014}. 

Andrew E et al. proposed that the spin image feature (Spin Image) is a local point cloud feature extraction method based on image processing\cite{johnson_using_1999}. The basic idea is to convert the point cloud into a two-dimensional grayscale image and then perform statistics on it. processing and analysis. In the specific implementation, each point in the point cloud is regarded as a central point, the distribution of the nearest point is calculated, and the information is encoded into a vector of uniform length to describe the characteristics of the point cloud. The SHOT (Signature of Histograms of Orientations) feature is a local point cloud feature extraction method proposed by Tombari et al\cite{salti_shot_2014}. It extracts point cloud features by calculating the relative position and normal vector of the surrounding points of a point in the point cloud. FPFH (Fast Point Feature Histogram) is also a local point cloud feature descriptor, which is optimized for the PFH (Point Feature Histogram) algorithm\cite{rusu_fast_2009}. In comparison to the PFH algorithm, the FPFH algorithm can greatly reduce the amount of calculation while maintaining accuracy and is suitable for real-time processing and large-scale point cloud data processing. The features of Spin Image, SHOT, and FPFH are all point cloud local feature extractions, which have good rotation invariance and scale invariance. Based on predecessors, RoPS and GASD descriptors further improve the rotation and scale invariance of local features through rotation statistics or the introduction of a fixed coordinate system\cite{guo_rotational_2013,silva_do_monte_lima_efficient_2016}. However, point cloud data generally does not only calculate the local features of a point, and usually needs to randomly sample multiple points and calculate their features. This leads to a large amount of calculation required for this type of algorithm in the process of practical application. However, point cloud global features can better reduce the amount of computation.

With the development of deep learning technology, neural networks for point cloud data processing have also been widely reported\cite{georgiou_survey_2020,qi_review_2021}. In order to learn the surface features of point cloud data, Charles R. Qi et al. proposed the PointNet neural network in 2017\cite{qi_pointnet_2017-1}. PointNet is an end-to-end deep learning model that can directly process point cloud data without converting point clouds into voxels, images, or extracting other types of features. The model processes each point in the point cloud through a fully connected neural network, and then extracts the feature representation of the point cloud through the pooling layer. Although PointNet is an end-to-end model that can quickly classify and segment point cloud data, this model does not consider the local characteristics of point clouds, so the recognition effect in some point clouds with complex results is not ideal. PointNet++ introduces a hierarchical feature extraction method based on PointNet\cite{qi_pointnet_2017}. Specifically, the PointNet++ model divides the point cloud data, then learns local features for different parts of the point cloud data, and then realizes the fusion of point cloud features in different regions through the fully connected layer. Therefore, compared with the former, the PointNet++ model has a better learning effect on the local features of point cloud data, and it also performs ideally in point cloud classification and segmentation tasks.

In recent years, there are also many models that combine the traditional features of point clouds with PointNet++ and have achieved good results in fields such as face recognition\cite{georgiou_survey_2020,cao_rp-net_2022}. But for the video retrieval task faced in this paper, the PointNet++ model itself has low computational complexity and simple structure, so it is more suitable as a video content feature extraction model.

\section{Marking Anything Algorithm Description}

\begin{figure}
  \centering
  \includegraphics{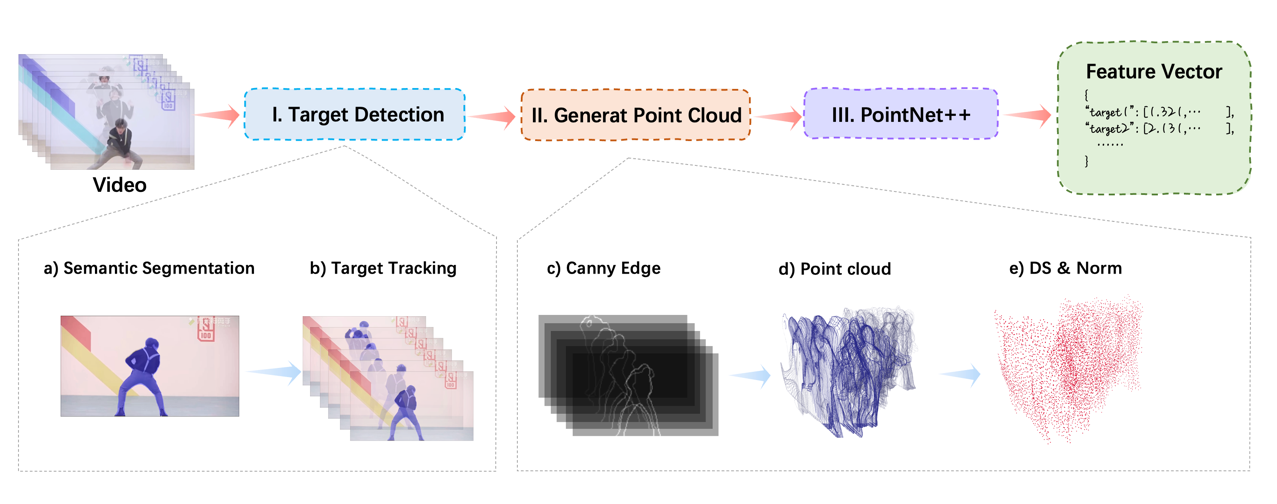}
  \caption{MA Algorithm Flowchart.}
  \label{fig:fig2}
\end{figure}

Fig.\ref{fig:fig2} shows the overall flow of the Marking anything algorithm. The algorithm model can divided into three modules, which are the target detection module shown in Fig.\ref{fig:fig2}-I, the point cloud generation module shown in Fig.\ref{fig:fig2}-II, and the point cloud feature extraction module shown in Fig.\ref{fig:fig2}-III.

The main function of the target detection module in Part I is to obtain the mask information of the detected target in each frame of the video, which includes the correspondence between the detected targets in the frame. As shown in Fig.\ref{fig:fig2}-a), the module loads the video at the input and obtains the video frame sequence $\{f_1,f_2,f_3,\ldots,f_n\}$. Module I first uses YOLO-v8 to perform target detection and pixel-level semantic segmentation frame by frame to obtain the mask information sequence of each detected target. The mask data of each frame is shown in Fig.\ref{fig:fig2}-a) blue mask. Then, as shown in Fig.\ref{fig:fig2}-b), module I uses the Deep-OC-SORT model to achieve the corresponding matching between multiple target frames, to obtain the motion vector information of the object between video frames. In summary, after being processed by module I, the mask sequences of all detected targets in the current video are obtained, as shown in Formula-(1), (2).

\begin{equation}
 Video \rightarrow \{C_1,C_2,C_3, \ldots,C_m\}
\label{eq:formula_1}
\end{equation}

\begin{equation}
C_i=\{mask_1,mask_2,mask_3,\ldots ,mask_n\}
\label{eq:formula_2}
\end{equation}

Among them, $C_i$ represents the $i_{th}$ target detected in the video, $mask_j\in\Re^{H\times W}$ represents the $j_{th}$ frame mask data map of the $i_{th}$ detected target, each element of which is a Boolean value, H and W respectively indicate the height and width of the mask image.

After being processed by module I, the mask information of each detected target in each frame of the video is output. Module II receives the information and processes it to obtain the point cloud outline of each target in the time dimension of the video. Specifically, for a single detection target, as shown in Fig.\ref{fig:fig2}-c), module II uses the Canny edge algorithm to extract the contour of a single target mask data $mask_j$, obtain a single target boundary map $canny_j$, and then obtain all the boundaries of the detection target Graph sequence                  $\{canny_1,canny_2,canny_3,\ldots,canny_n\}$,$canny_j\in\Re^{H\times W}$. Next, it is necessary to convert the boundary sequence information of the detection target $C_i$ into the three-dimensional coordinate data of the point cloud. The conversion relationship is shown in Formula-(3).
\begin{equation}
 Mask_{(H,W,N)} \rightarrow PCD_{(X,Y,Z)}
\label{eq:formula_3}
\end{equation}

Among them, $Mask$ represents the mask sequence coordinate system, and $H$, $W$, and $N$ represent the height, width, and sequence number of the mask image, respectively. PCD represents the point cloud data, and $X$, $Y$, and $Z$ represent the three coordinate systems of the point cloud, corresponding to the three coordinate systems of $H$, $W$, $N$, respectively.

The converted point cloud data is the spatial point cloud representation of a single detected object in the video, as shown in Fig.\ref{fig:fig2}-d). At this point, the point cloud expressions $                                \{PCD_1,PCD_2,PCD_3,\ldots,PCD_m\}$ of all detected targets are obtained. Because there are too many points in a single point cloud data, in order to simplify the subsequent calculation complexity, this paper uses the method of down sampling the farthest point and normalizing the coordinates to simplify the point cloud data, and obtain a new multi-object point cloud expression $\{PCD_{1}^{'},PCD_{2}^{'} ,PCD_{3}^{'},\ldots, PCD_{m}^{'}\}$. The simplified point cloud data is shown in Fig.\ref{fig:fig2}-e).

The PointNet++ network in Part III is mainly used to extract the surface feature expression vector of point cloud data. The model is mainly used for the classification and segmentation of point cloud data. In order to adapt to the goal of extracting the contour feature vector of point cloud data, this paper uses the Modelnet40 data set to pre-train the model for classification tasks. The output vector of the last fully connected layer of the model is used as the expression vector of the point cloud shape feature, which is obtained from $\{PCD_{1}^{'},PCD_{2}^{'},PCD_{3}^{'},\ldots, PCD_{m}^{'}\}$ through the PointNet++ network $\{feature_1,feature_2,feature_3,\ldots,feature_m \}, feature_i\in\Re^{1\times 256}$.

As mentioned above, after the processing of the three modules, the contour feature expression of all detection targets in the video is obtained, and the expression vector is the retrieval basis of the video detection target.

\section{Experiments}
\subsection{A simple searchable video database}

\begin{figure}
  \centering
  \includegraphics{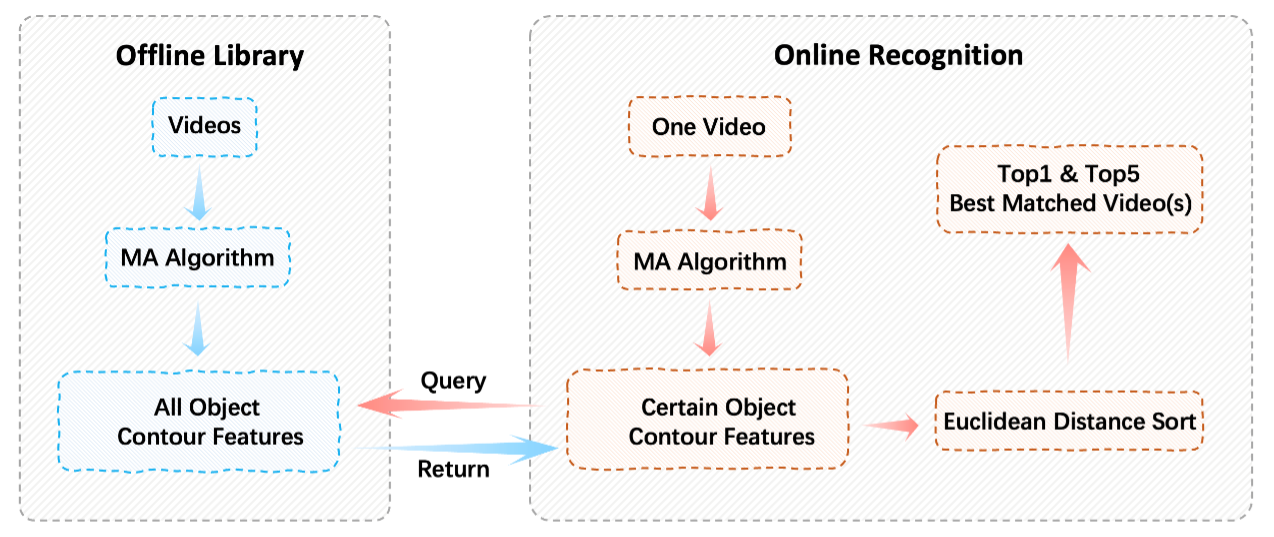}
  \caption{A searchable video database flow chart.}
  \label{fig:fig3}
\end{figure}

Before starting to verify the validity of contour features, it is necessary to design the retrieval process and validity verification parameters of video, that is, to design several simple and retrievable video databases. As shown in Fig.\ref{fig:fig3}, the matching process for a detection target needs to be divided into two parts, namely building an offline database and online target recognition. Specifically, the offline database uses the MA algorithm proposed in this paper to extract the contour features of multiple targets from all the collected videos and store them in the database. Online target recognition is to extract the contour features of a single target in a single video, and then use this feature to calculate the corresponding Euclidean distance with the features of all targets in the offline database. Then sort the calculated Euclidean distance in ascending order and return the previous group or the first five groups of video ID numbers that match the most. These videos contain similar detection targets to the detected video.

The retrieval accuracy of the experimental data set design in this paper refers to the ratio of the correctly retrieved videos in top1 or top5 in the data set to the total number of data sets.

\subsection{The performance of contour feature on video retrieval}

In order to verify the effectiveness of detecting the contour features of objects, this paper selects video clips with high content similarity from DouYin. The detected targets and video numbers of the three datasets are shown in Tab.\ref{tab:tab1}. The detection targets are people, cats, and dogs. The screenshots of different video clips are shown in Fig.\ref{fig:fig1} a)-b). In these video clips, the silhouettes, actions, and positions of people, cats, and dogs have high Similarity, so these videos can be used to test the effectiveness of the contour features proposed in this paper for extracting features from similar video content.

\begin{figure}
  \centering
  \includegraphics{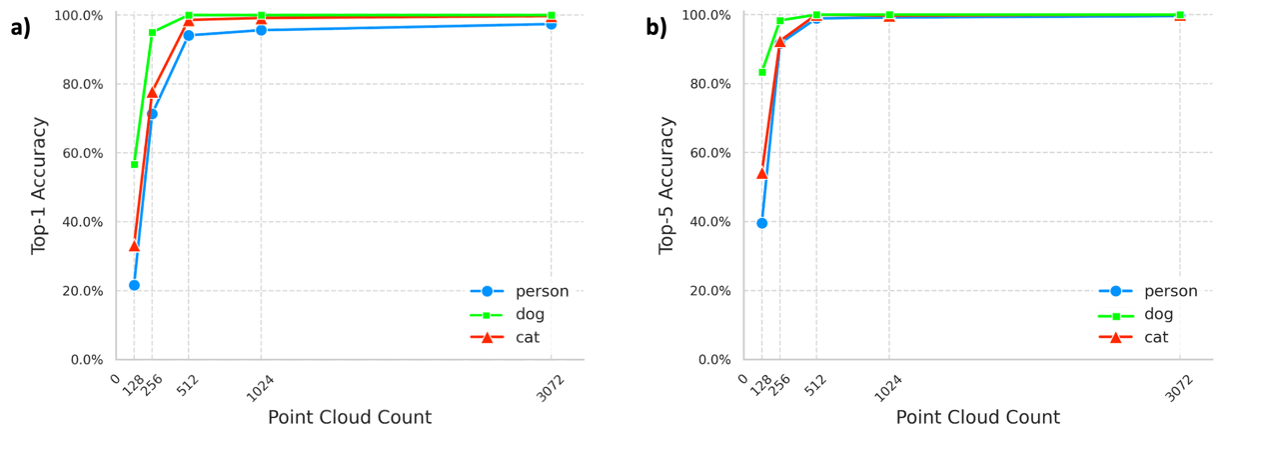}
  \caption{Influence of the point count of point cloud on the accuracy of contour feature retrieval, a) Top1 best matching accuracy, b) Top5 best matching accuracy.}
  \label{fig:fig4}
\end{figure}

\begin{table}
 \centering
 \caption{DouYin dataset}
 \begin{tabular}{cccc}
   \toprule
   Target Type & Person & Cat & Dog \\
   \midrule
   Number & 730 & 356 & 60 \\
   \bottomrule
 \end{tabular}
 \label{tab:tab1}
\end{table}
Use the experimental data set retrieval process proposed in 4.1 to extract the contour features of the corresponding detection target from the above video data. In order to verify the impact of point cloud point count on retrieval accuracy, this paper retains different points for each category of point cloud when down sampling, which are 128 points, 256 points, 512 points, 1024 points, and 3072 points. Note that because down sampling selects the method of sampling the farthest point, there is a certain degree of randomness. Therefore, according to the retrieval method in 4.1, the process of online recognition and offline database down sampling for the point cloud of the same target is required twice. Finally, the retrieval accuracy of Top1 and Top5 three types of target videos is shown in Fig.\ref{fig:fig4}.

As shown in Fig.\ref{fig:fig4}, whether it is the Top1 or Top5 retrieval method, the overall retrieval accuracy increases with the increase of the number of retained sampling points. Among them, when the number of points reaches 3072 points, the retrieval accuracy of the three types of objects reaches the maximum, and the specific accuracy is shown in Tab.\ref{tab:tab2}. Among the collected videos, the similarity between person-related videos is greater than that of cat and dog videos. However, when this type of video has 3072 sampling points, the Top1 accuracy reaches 97.4\%, and the Top5 accuracy reaches 99.6\%. The retrieval accuracy of cat and dog videos is also close to 100.0\% under the Top1 retrieval method. The above data show that the video object contour feature proposed in this paper is effective for constructing video content retrievable features. At the same time, the number of sampling points is 3072 points, which can ensure better retrieval accuracy.

\begin{table}
 \centering
 \caption{Contour feature retrieval accuracy of 3072 point cloud count}
 \begin{tabular}{cccc}
   \toprule
   Target Type & Person & Cat & Dog \\
   \midrule
   Top1 & 97.4\%	& 99.7\%	      & 100.0\% \\
   Top5 & 99.6\%	& 100.0\%	& 100.0\% \\
   \bottomrule
 \end{tabular}
 \label{tab:tab2}
\end{table}

\subsection{The performance of contour feature on edited video retrieval}

In order to further verify the effectiveness of the contour feature and expand the application scenarios of this feature. This paper randomly selects 1000 groups of videos containing humans from the Kinetics-700 dataset, and edits these videos, including changing the video aspect ratio, flipping the frame order of the video playback, cropping the video length, mirroring the video, and cropping the video to obtain a partial screen, play video at 2x speed, play video at 0.5x speed, and rotate video 90 degrees clockwise. In order to simplify the verification process, this paper selects the person with the longest appearance time as the only retrieval target for a video with multiple persons.

According to the MA algorithm proposed in this paper, the above-mentioned original video and the edited video are extracted from the contour features of the persons in the video for traceability retrieval. In order to verify the validity and robustness of the retrieval features of the contour features proposed in this paper after the video is edited. The retrieval accuracy of Top1 and Top5 is shown in Tab.\ref{tab:tab3}. For videos with changed aspect ratios and videos played in reverse order, the features proposed in this paper still have certain retrieval significance in the Top1 retrieval mode. In the Top5 retrieval mode, videos edited along the time dimension and mirrored videos also have retrieval significance.

However, the features proposed in this paper do not have effective practical value for retrieval for videos played at variable speed and rotated videos. This is because the MA algorithm does not have good rotation invariance for the features extracted from the point cloud in the PointNet++ part.

It is worth noting that for the retrieval of unedited videos, the Top1 retrieval accuracy reaches 99.7\%, which is higher than the Top1 retrieval accuracy of 97.4\% for the person videos in Section 4.2. This is because the contours, motions and other features of the targets in the person videos extracted by the Kinetics dataset are not similar, and the videos selected in Section 4.2 are high similar.

\begin{table}
 \centering
 \caption{Contour feature retrieval accuracy of 3072 point cloud count}
 \begin{tabular}{ccc}
   \toprule
     & Top1 & Top5\\
   \midrule
    Original Video	&	99.70\%	&	100.00\%	\\
    Change L/W of Video	&	88.80\%	&	99.50\%	\\
    Reverse Video	&	60.20\%	&	85.50\%	\\
    Clip Video	&	59.60\%	&	86.90\%	\\
    Mirrored video	&	47.60\%	&	73.90\%	\\
    Clip L\&W of Video	&	26.30\%	&	58.50\%	\\
    Video 2x speed	&	17.80\%	&	51.20\%	\\
    Video 0.5x speed	&	13.40\%	&	44.50\%	\\
    Video Rotation	&	2.30\%	&	10.20\%	\\
   \bottomrule
 \end{tabular}
 \label{tab:tab3}
\end{table}

\section{Conclusion}
This paper proposes an algorithm for marking anything in the video, which can extract the contour features of any target in the video and convert it into a feature vector with a length of 256 that can be retrieved through the point cloud feature extraction model PointNet++. At the same time, this paper selects some videos with high content similarity and deliberately edited videos and constructs a simple and retrievable video database through the MA algorithm, which verifies that the contour features proposed in this paper have high effectiveness and robustness. 

The contour feature extraction method proposed in this paper can be used to build a video database to manage and trace the video. In practical application scenarios, tracing the source of similar videos can not only mark the video with unique id information, but also be used to trace the source of false videos, which has quickly realized the rumor refusing. However, the current PointNet++ does not have a good rotation invariance, so the traceability of the rotated video is not ideal. In the future, we will continue to further improve the robustness of the algorithm by adding features with rotation invariance or strengthening the training data.

% \section*{Acknowledgments}
% This was was supported in part by......

%Bibliography
\bibliographystyle{unsrt}  
\bibliography{marking_anything}

\end{document}